\documentclass[sigconf]{acmart}

\copyrightyear{2020}
\acmYear{2020}
\setcopyright{acmcopyright}\acmConference[KDD '20]{Proceedings of the 26th ACM SIGKDD Conference on Knowledge Discovery and Data Mining}{August 23--27, 2020}{Virtual Event, CA, USA}
\acmBooktitle{Proceedings of the 26th ACM SIGKDD Conference on Knowledge Discovery and Data Mining (KDD '20), August 23--27, 2020, Virtual Event, CA, USA}
\acmPrice{15.00}
\acmDOI{10.1145/3394486.3403244}
\acmISBN{978-1-4503-7998-4/20/08}

\usepackage{amssymb}

\usepackage{latexsym}

\usepackage{url}

\usepackage{multirow}
\usepackage{balance}
\usepackage{amsfonts}
\usepackage{amsmath}
\usepackage{bbold}

\usepackage{amsthm}
\usepackage{graphicx}

\usepackage{microtype}
\usepackage{url}
\usepackage{enumitem}
\usepackage{wrapfig,lipsum}
\usepackage{multirow}
\usepackage{makecell}
\usepackage{wrapfig}
\usepackage{tabularx}
\usepackage[boxed, ruled,vlined,linesnumbered]{algorithm2e}
\usepackage{algpseudocode}
\usepackage{microtype}
\usepackage{etoolbox}
\usepackage{subcaption}
\usepackage{float}
\usepackage{booktabs}
\usepackage{caption}
\usepackage{blindtext}
\usepackage[linesnumbered,ruled]{algorithm2e}%

\usepackage{pifont}
\usepackage{tabularx,ragged2e}

\usepackage{environ}

\newcommand{\jh}[1]{}
\newcommand{\la}{\mbox{$\langle$}}
\newcommand{\ra}{\mbox{$\rangle$}}

\usepackage{geometry} \usepackage{array, makecell}%

\newcommand{\nop}[1]{}
\newcommand{\bs}[1]{\boldsymbol{#1}}

\newcommand{\corel}{\mbox{\sf CoRel}\xspace}

\newcommand{\wrong}[1]{#1 ($\times$)}
\newcommand{\redund}{\ding{34}}
\newcommand{\pc}{$\langle$$p$,$c$$\ra$ }
\newcommand{\pczero}{$\langle$$p_0$,$c_0$$\ra$ }

\AtBeginDocument{%
  \providecommand\BibTeX{{%
    \normalfont B\kern-0.5em{\scshape i\kern-0.25em b}\kern-0.8em\TeX}}}

\setcopyright{acmcopyright}
\begin{document}

\fancyhead{}
\leftmargini=12pt
\title{CoRel: Seed-Guided Topical Taxonomy Construction\\ by Concept Learning and Relation Transferring}




 \author{Jiaxin Huang$^{1}$, Yiqing Xie$^{2}$, Yu Meng$^{1}$, Yunyi Zhang$^{1}$, Jiawei Han$^{1}$}
 \affiliation{
 \institution{$^1$University of Illinois at Urbana-Champaign, IL, USA} 
 \institution{$^2$The Hong Kong University of Science and Technology, Hong Kong, China}
 \institution{$^{1}$\{jiaxinh3, yumeng5, yzhan238, hanj\}@illinois.edu \ \ \ $^2$yxieal@ust.hk}
 }


\begin{abstract}
Taxonomy is not only a fundamental form of knowledge representation, but also crucial to vast knowledge-rich applications, such as question answering and web search. 
Most existing taxonomy construction methods extract hypernym-hyponym entity pairs to organize a ``universal'' taxonomy. 
However, these generic taxonomies cannot satisfy user's specific interest in certain areas and relations. 
Moreover, the nature of instance taxonomy treats each node as a single word, which has low semantic coverage\jh{for people to fully understand}. 
In this paper, we propose a method for seed-guided topical taxonomy construction, which takes a corpus and a seed taxonomy described by concept names as input, and constructs a more complete taxonomy based on user's interest, wherein each node is represented by a cluster of coherent terms. 
Our framework, \corel, has two modules to fulfill this goal. 
A relation transferring module learns and transfers the user's interested relation along multiple paths to expand the seed taxonomy structure in width and depth. 
A concept learning module enriches the semantics of each concept node by jointly embedding the taxonomy and text. 
Comprehensive experiments conducted on real-world datasets show that \corel generates high-quality topical taxonomies and outperforms all the baselines significantly.
\end{abstract}



\maketitle

%
%
%
%
\section{Introduction}

\begin{figure}[t]
\centering
\includegraphics[width=1\linewidth]{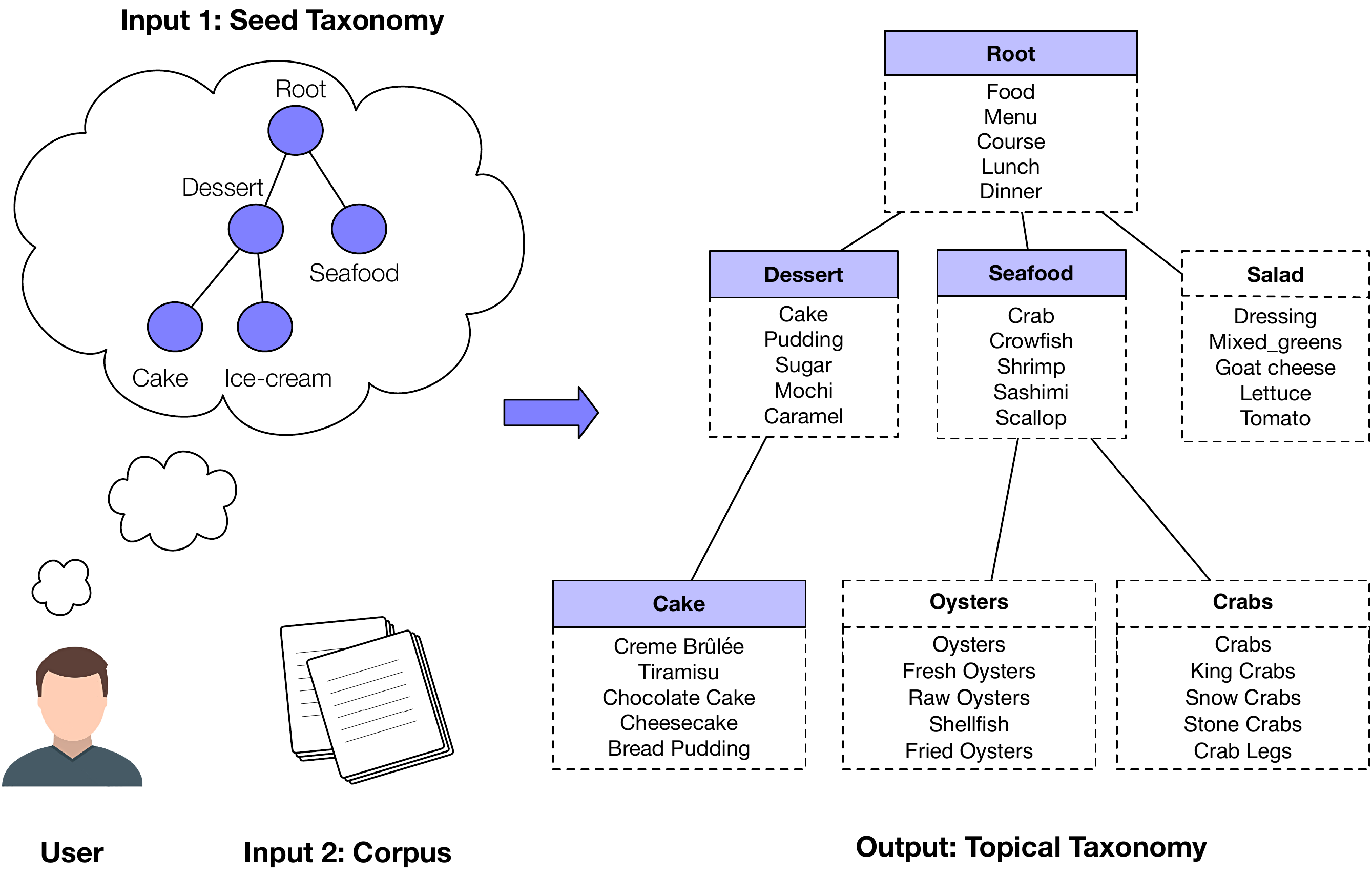}
\caption{Seed-guided topical taxonomy construction. User inputs a partial taxonomy, and \corel extracts a more complete topical taxonomy based on user-interested aspect and relation, with each node represented by a cluster of words.}
\label{fig:output}
\vspace{-0.5cm}
\end{figure}

Taxonomy is an essential form of knowledge representation and plays an important role in a wide range of applications \cite{Zhang2014TaxonomyDF,Yang2017EfficientlyAT,Meng2020HierarchicalTM}. 
A taxonomy constructed from a large corpus organizes a set of concepts into a hierarchy, making it clear for people to understand relations between concepts. 

Most existing taxonomy construction methods organize hypernym-hyponym entity pairs into a tree structure to form an instance taxonomy. However, a ``universal'' taxonomy so constructed cannot cater to user's specific needs.
For example, a user might want to learn about concepts in a certain aspect (e.g., \textit{food} or \textit{research areas}) from a corpus.
Generic taxonomy has two noteworthy limitations: 
(1) Countless irrelevant terms and fixed ``is-a'' relations dominate the instance taxonomy, failing to capture user's interested aspects and relations, and
(2) each node is represented by a single word without considering term correlation, limiting people's understanding due to low semantic coverage, not to mention that synonyms could appear at multiple nodes.

We study the problem of seed-guided topical taxonomy construction, where user provides a seed taxonomy as guidance, and a more complete topical taxonomy is generated from text corpus, with each node represented by a cluster of terms (topics). 
As shown in Figure \ref{fig:output}, a user provides a seed taxonomy and wants to generate a more complete food taxonomy from a given corpus.
Such a more complete topical taxonomy can be hopefully constructed by expanding various types of food both in width and depth, with a cluster of descriptive terms for each concept node as a topic. 

To fulfill this, we propose a framework \corel, which approaches the problem with two modules:
(i)
A relation transferring module learns the specific relation preserved in seed taxonomy and attaches new concepts to existing nodes to complete the taxonomy structure.;
and (ii)
a concept learning module captures user-interested aspects and enriches the semantics of each concept node. 
Two challenges are met in the course: 
(1) Fine-grained concept names can be close in the embedding space, and enriching the concepts might result in relevant but not distinctive terms (e.g., ``sugar'' is relevant to both ``cake'' and ``ice-cream''); and 
(2) with minimal user input, it is nontrivial to directly apply weakly supervised relation extraction methods to expand the taxonomy structure.
Overall, noisy terms may harm the quality of new topics found.

To address these challenges, the relation transferring module first captures the relation preserved in seed parent-child pairs and transfers it upwards and downwards for finding the first-layer topics and subtopics, attached by a co-clustering technique to remove inconsistent subtopics.
The concept learning module learns a discriminative embedding space by jointly embedding the taxonomy with text and separating close concepts in the embedding space. 

We demonstrate the effectiveness of our framework through a series of experiments on two real-world datasets, and show that \corel outperforms all the baseline methods in multiple metrics. 
We also provide qualitative analysis to demonstrate the high quality of our generated topical taxonomy and its advantages over other methods.

Our contribution can be summarized as follows: 
(1) A novel framework for seed-guided topical taxonomy construction.
(2) A relation transferring module that passes the user-interested relation along multiple paths in different directions for taxonomy structure completion.  
(3) A concept learning module that enriches the semantics for a taxonomy of words by extracting distinctive terms. 
(4) Comprehensive experiments on real-world data with qualitative and quantitative studies that prove the effectiveness of \corel.

\vspace{-0.1cm}
\begin{figure*}[t]
\centering
\includegraphics[width=1.0\linewidth]{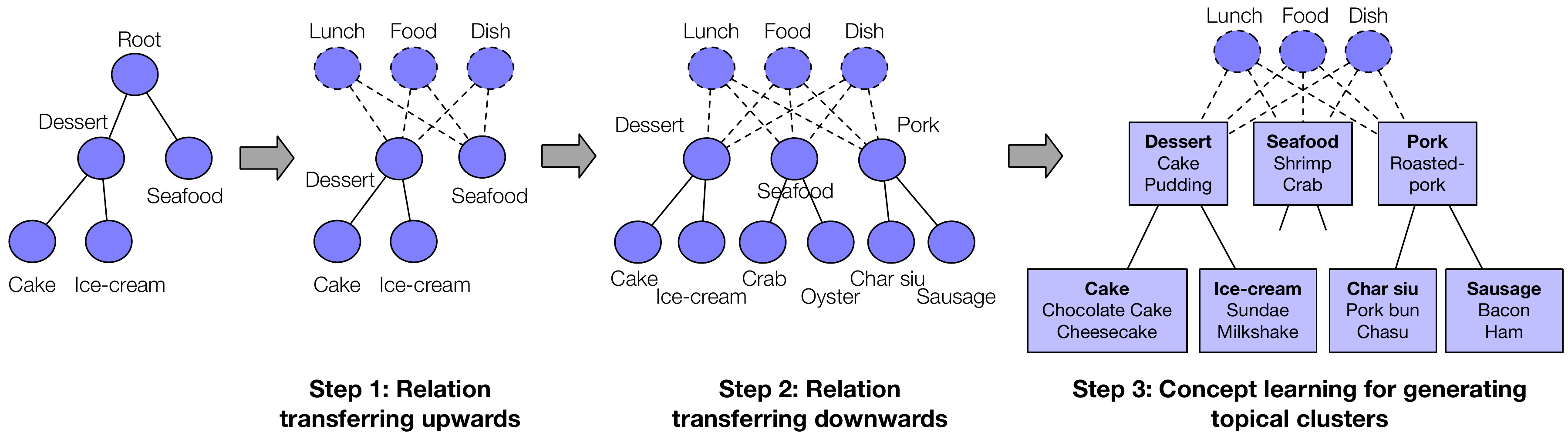}
\caption{Workflow of \corel. }
\label{fig:workflow}	
\vspace{-0.3cm}
\end{figure*}

\section{related work}

\smallskip
\noindent \textbf{Unsupervised Taxonomy Construction.}~

Most existing taxonomy construction algorithms perform a two-step approach: Hypernym-hyponym pairs are first extracted from a corpus and then organized into a tree structure.
The task of hypernym-hyponym extraction traditionally relies on pattern-based methods which utilize Hearst patterns~\cite{Hearst1992AutomaticAO} like ``NP such as NP'' to acquire parent-child pairs that satisfy the ``is-a'' relation. Later, researchers design more lexical patterns~\cite{Navigli2010LearningWL,Nakashole2012PATTYAT} or extract such patterns automatically~\cite{Snow2004LearningSP,Shwartz2016ImprovingHD} in a bootstrapping method. 
These pattern-based methods suffer from low recall due to the diversity of expressions.
Distributional methods alleviate the problem of sparsity by representing each word as a low-dimensional vector to capture their semantic similarity. There exist approaches~\cite{Weeds2004CharacterisingMO, Chang2017DistributionalIV} inspired by Distributional Inclusion Hypothesis~\cite{ZhitomirskyGeffet2005TheDI} (the context of a hyponym should be the subset of that of a hypernym) to detect hypernym-hyponym pairs without supervision. 
On the end of term organization, graph-based methods~\cite{Kozareva2010ASM,Navigli2011AGA,Velardi2013OntoLearnRA} are used to remove conflicts that form loops in the taxonomy. 
The aforementioned algorithms are not suitable for constructing a topical taxonomy, since they treat each word as a single node instead of forming topics with relevant terms for people to comprehend.
 
As another line of work, clustering-based taxonomy construction is closer to our problem setting.  Clustering-based taxonomy construction methods first learn a representation space for terms, then perform clustering to separate terms into different topics by different measures~\cite{Cimiano2004ComparingCD, Liu2012AutomaticTC, Wang2013APM,Yang2009AMF}. 
A recent method TaxoGen~\cite{Zhang2018TaxoGenUT} finds fine-grained topics by spherical clustering and local-corpus embedding. 
Hierarchical topic modeling algorithms~\cite{Blei2003HierarchicalTM,Mimno2007MixturesOH} are comparable to these methods, since they organize terms to form a taxonomy of topics, each represented by a word distribution.

The above methods do not require supervision in taxonomy construction.
They suffer from two disadvantages: (1) Without user input seed terms, they cannot capture users' specific interest in certain aspects of the corpus (e.g., ``food'' or ``research areas''), thus the final output may include a large number of irrelevant terms; and (2) these methods either capture generic ``is-a'' patterns or do not enforce specific relations in children finding (clustering-based methods), thus cannot cater to user-interested relations.

\smallskip
\noindent \textbf{Seed-Guided Taxonomy Construction.}~

For seed-guided taxonomy construction, HiExpan~\cite{Shen2018HiExpanTT} integrates the above two-step approach into a tree expansion process by width and depth expansion of the original seed taxonomy. Specifically, for width expansion that adds sibling nodes to those sharing the same parent, the method uses a set expansion algorithm~\cite{Shen2017SetExpanCS} that leverages skip-gram features to calculate similarity between terms. For depth expansion that attaches children nodes to new node (e.g., attaching ``oyster'' to ``seafood'' in Figure~\ref{fig:output}), they use word analogy~\cite{Mikolov2013DistributedRO} to capture relations between parent-child pairs. 
However, in our setting of constructing topical taxonomy, HiExpan suffers from two drawbacks: (1) the set expansion algorithm is not good at expanding concepts; and (2) word analogy is only locally preserved in the Word2Vec space~\cite{Fu2014LearningSH}. 

\smallskip
\noindent \textbf{Weakly Supervised Relation Extraction.}~

To construct a taxonomy that fits in with a user-interested relation, we aim to preserve the same relation between all newly added parent-child topics. With only a few given seeds, it is impossible to train a highly accurate and complicated relation extraction model with a huge number of parameters. 
Traditional weakly supervised relation extraction methods~\cite{Snow2004LearningSP,Nakashole2012PATTYAT} find textual patterns from given instances, suffering from sparsity of relation expressions. Recent studies~\cite{Qu2017WeaklysupervisedRE} combine textual patterns with distributional features for mutual enhancement in a co-training framework. 
Neural-based methods like prototypical network~\cite{Gao2019HybridAP} which represents each relation as a vector have shown to be effective in few shot relation (FSL) extraction. Recent advances in contextualized text representation show that deep transformers (e.g., BERT~\cite{Devlin2019BERTPO}) learn task-agnostic representations achieving strong performance on various NLP tasks. Researchers show that by learning from large amounts of entity pairs co-occurring in Wikipedia corpus, BERT can achieve state-of-the-art~\cite{Soares2019MatchingTB} on FSL relation extraction on benchmark datasets~\cite{Han2018FewRelAL}.

\smallskip
\noindent \textbf{Supervised Taxonomy Construction.}~

Most supervised taxonomy construction methods focus on extracting hypernym-hyponym pairs. Word analogy $(v(\text{man}) - v(\text{king})$ $=v(\text{woman})-v(\text{queen})$) is preserved in local clusters, and a piecewise linear projection from words to their hypernyms is trained in~\cite{Fu2014LearningSH}. For neural-based methods, order embedding \cite{Vendrov2015OrderEmbeddingsOI,Dash2019HypernymDU} is proposed to express partial orders between words. Later studies show that Poincar\'e space can be viewed as a continuous generalization of tree structures, so they embed large taxonomy structures extracted from WordNet~\cite{Nickel2017PoincarEF} or Wikipedia~\cite{Le2019InferringCH}. 
However, in our setting, the user-given taxonomy is of very limited size, and thus cannot be adaptive to these frameworks.

\section{Problem Description}
In this section we describe the task of seed-guided topical taxonomy construction.
 The inputs are a collection of documents $\mathcal{D}=\{d_1,d_2,...,d_{|\mathcal{D}|}\}$ and a tree-structured seed taxonomy $T^0$ provided by user. Each node $e$ in $T^0$ is represented by a single word from the corpus, and each edge $\la$$p_0$,$c_0$$\ra$ implies user-interested relation between a parent-child pair, such as ``is a subfield of'' or ``is a type of''. The output is a more complete topical taxonomy $T$, with each node $e$ as a conceptual topic, represented by a coherent cluster of words describing the topic. Figure \ref{fig:output} shows an example of our task.
 
The meanings of the notations used in this paper  are presented in Table \ref{tab:not}.
 
\begin{table}[ht]
\caption{Notations and Meanings.}
\label{tab:not}
\scalebox{1.0}{
\begin{tabular}{cc}
\toprule
Notation & Meaning \\
\midrule
$T$ & The tree structure of taxonomy. \\
$R$ & Root node of taxonomy $T$. \\
$e$ & A concept node on taxonomy $T$. \\
$C_e$ & The topic cluster of concept $e$. \\
$N_e$ & Nodes sharing common parent with $e$, including $e$. \\
$B_e$ & The children nodes of $e$. \\
\pc & A pair of direct parent and child nodes in $T$.\\
\bottomrule
\end{tabular}
}
\end{table}

\section{Methodology}

In this section, we introduce our proposed method by first giving an overview in Section \ref{sec:overview} and then describing the details of two modules in Sections \ref{sec:gen} and \ref{sec:exp}.

\subsection{Method Overview}\label{sec:overview}
Figure \ref{fig:workflow} shows the workflow of \corel. To expand the tree structure of a user-given seed taxonomy $T^0$, \corel first leverages a relation transferring module to capture seed relations of edge $\la$$p_0$,$c_0$$\ra$. In step 1, it attempts to discover potential root concepts by transferring the relation upwards, such as ``Lunch'', ``Food'' and ``Dish'' as more general concepts to cover the topics. In step 2, the relation is transferred downwards to attach new topics (internal nodes) as well as new subtopics (leaf nodes). Finally, a concept learning module is used to learn a discriminative embedding space to generate topical clusters for each concept node. Below we address the two modules in detail.

%


\subsection{Taxonomy Completion by Relation Transferring}\label{sec:exp}
The relation transferring module is used to complete the taxonomy structure by finding new topics and subtopics.
This module first captures the relation between user given \pc pairs by training a relation classifier on the given corpus. The relation classifier takes a relation statement (will be defined in section \ref{sec:rel}) of a pair of terms as input, and judges whether there exists user-interested relation and which direction it is between the pair. 
After training the relation classifier, we transfer the relation upwards for root node discovery, and then transfer the relation downwards to find new topics/subtopics as the child of root/topic node. In both the example and evaluation we construct two layers of topics, though this module can be applied to discover more fine-grained topics by further going down.

\subsubsection{Self-supervised Relation Learning}\label{sec:rel}
\begin{figure}[t]
\centering
\includegraphics[width=1\linewidth]{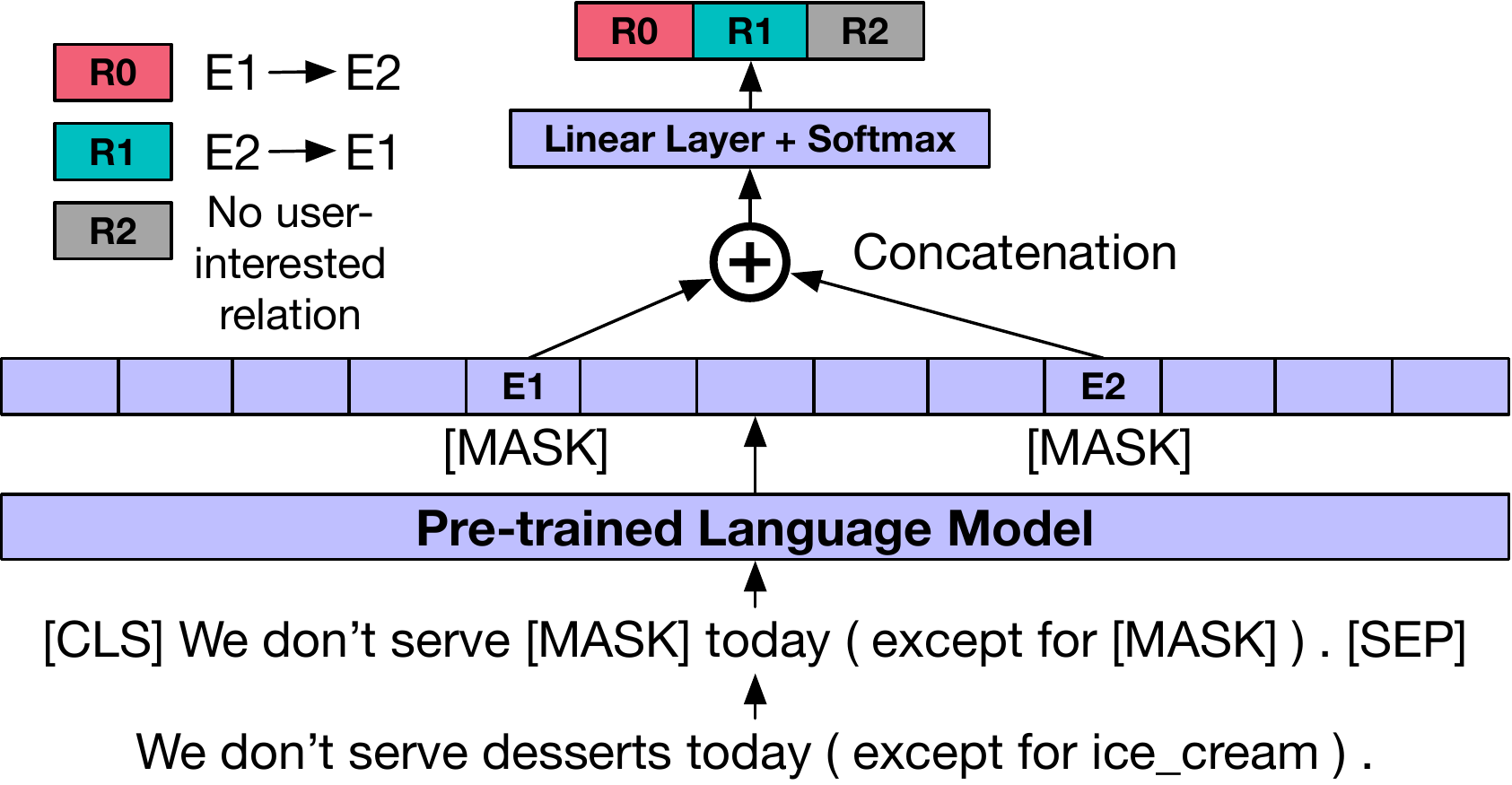}
\caption{Our Weakly Supervised Relation Classifier.}
\label{fig:bert}
\end{figure}

 Previous studies show that word analogy can capture relations between words to some extent \cite{Mikolov2013DistributedRO, Shen2018HiExpanTT}, but vector offset is only preserved in a local area in the embedding space \cite{Fu2014LearningSH}. To deal with more complicated relations, our choice of the model is inspired by the effectiveness of the pre-trained deep language model, BERT\cite{Devlin2019BERTPO}, on wide downstream applications. \cite{Soares2019MatchingTB} also shows its power in few-shot relation learning by training the model in a distant supervised setting using large amounts of pairs of entities on Wikipedia corpus. In our setting, user gives minimal seeds which limits the potential to train such deep language model, thus we only employ the pre-trained BERT model and train a relation classifier as shown in Figure \ref{fig:bert}. 

\smallskip
\noindent \textbf{Relation Statement.}~
We assume that if a pair of \pc co-occurs in a sentence in the corpus, then that sentence implies their relation. We refer to sentences containing \pc as their relation statements and leverage a pre-trained deep language model to understand the relation statements. To learn the user-interested specific relation, we extract all the relation statements of user-given \pczero as positive training samples. We collect negative training sentences in two ways: (1) relation statements of sibling nodes, thus avoiding the model to only find closely related terms; and (2) random sentences from the corpus, so the model can learn from irrelevant contexts to avoid overfitting.

\smallskip
\noindent \textbf{Sequence Input Representation.}~
Since user gives only a minimal number of seeds, which is not enough to train deep encoders, we cannot simply add explicit markers around pairs of terms to let the model pay attention to. Therefore, we take the original sentence and replace the two terms by ``[MASK]'' tokens with two justifications: (1) aligning with the pre-trained objective of masked language model; and (2) avoiding the classification layer to only remember relations from training pairs instead of looking into contexts.

\smallskip
\noindent \textbf{Classification Layer.}~
We take the output of two ``[MASK]'' tokens from the last layer of the pre-trained language model, and concatenate them to be the input of the classification layer, where we use a simple linear layer before the softmax layer. The output label chooses the relation from $e_1$ to $e_2$ among three classes in the relation set $\mathcal{Q}$: $\la$$e_1$$\rightarrow$$e_2$$\ra$ (i.e., $e_1$ is a parent node of $e_2$), $\la$$e_2$$\rightarrow$$e_1$$\ra$, or non-user interested relation.

\smallskip
\noindent \textbf{Data Augmentation.}~
To fully utilize the asymmetric property along the taxonomy edges, we augment each input training sequence by reversing the order of concatenation of $e1$ and $e2$. Then the label would switch if user-interested relation exists between the pair, but will not change otherwise.

\subsubsection{First-layer Topic Finding by Root Node Discovery}\label{sec:width}
After deriving a relation classifier, we can easily transfer the user-interested relation along the paths in the taxonomy. This is done by targeting an existing node and finding entities to be its potential parent node (transferring upwards) or child node (transferring downwards).
To ``expand'' more first-layer topics based on user-given ones, previous work like set expansion is good at extending a list of instances like country names or company names \cite{Shen2017SetExpanCS, Rong2016EgoSetEW,Huang2020GuidingCS}, but is not perfect at expanding concept names. Another seemingly straightforward solution is to train a few-shot sibling relation classifier based on relation statements of sibling nodes. However, this would result in low recall of extracted new topics, since the co-occurrence of two relatively unrelated topics might be sparse in the corpus.

To resolve the above issue, we assume that if we can discover potential root nodes, such as ``Food'' for ``Dessert'' and ``Seafood'', then the root node would have more general contexts for us to find connections with potential new topics. Previous studies \cite{Meng2020DiscriminativeTM, ZhitomirskyGeffet2005TheDI} also found general concepts cover broader semantic meaning and have more varied contexts.

Specifically, we transfer the relation upwards by using the relation classifier learned to extract a list of parent nodes for each seed topic. The common parent nodes for all topics are treated as root nodes $R$. 

\smallskip
\noindent \textbf{Finding common root nodes.}~
To find the parent or child of an existing node, we extract relation statements of a concept $e$ and a candidate term $w$ into the relation classifier to judge sentence-based relations. Corpus-based relation between $w$ and $e$ is then averaged over confident sentence-based results over the corpus, with the confidence threshold being $\delta$.
\begin{align}\label{eq:score}
\text{Score}(w \rightarrow e) &= \frac{\sum_{s_{w\rightarrow e}}  \mathbb{1} \left(KL\left(\bs{l} \middle\| \bs{p}_{w} \right) > \delta \right)}{\sum_{q\in \mathcal{Q}} \sum_{s_q} \mathbb{1} \left(KL\left(\bs{l} \middle\| \bs{p}_{w} \right) > \delta \right) }
\end{align}
where $s_q$ denotes relation statements in which the relation of $q$ exists, $\bs{p}_{w}$ denotes the output probability from the relation classifier, and $l$ is the uniform distribution vector among three classes of relations. Thus if the KL divergence between the two distributions is larger than a threshold $\delta$, we treat the prediction as a confident one. Eq.~\eqref{eq:score} calculates the portion of term $w$ being the parent of concept $e$ among all the confident predictions, and we confirm $w$ as the parent node of $e$ if the portion is larger than a threshold. 
For each user-given first-layer topic, we can generate a list of parent nodes, and their common parent nodes are treated as root nodes $R$. 

\smallskip
\noindent \textbf{Finding new first-layer topics.}~
We apply the relation classifier to extract child terms for each root node $r \in R$. This is done in a similar way as root node discovery, but we only reverse the direction of relation. Thus we need to replace $(w \rightarrow e)$ in Eq.~\eqref{eq:score} with $(r \rightarrow w)$. New topics are selected by their average score over all root nodes.
\begin{align}\label{eq:extraScore}
\text{Score}(R\rightarrow w)=\frac{\sum_{r \in R}\text{Score}(r\rightarrow w)}{|R|}
\end{align}



\subsubsection{Candidate term extraction for subtopics}\label{sec:cand}
After generating the first-layer topics, we transfer the relation downwards to discover subtopics of each first-layer topic. This can be done by applying Eq.~\eqref{eq:score} again and replacing $(w \rightarrow e)$ with $(e \rightarrow w)$. The candidate terms will later be clustered into subtopics.

\subsection{Generating Topical Clusters by Concept Learning}\label{sec:gen}

Our concept learning module is used to learn a discriminative embedding space, so that each concept is surrounded by its representative terms. Within this embedding space, subtopic candidates are also clustered to form coherent subtopic nodes. This is motivated by the fact that relevant concept names can be close to each other in the embedding space, and directly using unsupervised word embedding such as Word2Vec \cite{Mikolov2013DistributedRO} might result in relevant but not distinctive terms (e.g., ``food‘’ is relevant to both ``seafood‘’ and ``dessert‘’). Thus we use the expanded taxonomy as input to guide the word embedding learning process.


\subsubsection{Concept Learning based on Taxonomy and Corpus}\label{sec:emb}
We basically design three loss functions to embed the concepts, words and documents in a joint embedding space.
Our starting point is the assumption that similar words share similar local contexts, as is used by the Skip-Gram model \cite{Mikolov2013DistributedRO}. Two sets of embedding for each word are used: center word embedding denoted as $u_w$ and context word embedding as $v_w$. The training objective is to maximize the log probability of observing words in a fixed local context window with the size of h.
\begin{equation}
    \label{eq:local}
    \begin{aligned}
        L_l &= -\sum_{d\in D} \sum_{1\leq i\leq |d|} \sum_{0<|j-i|\leq h} \log P(w_j\mid w_i) 
    \end{aligned}
\end{equation}
where $P(w_j\mid w_i) \propto \exp{\left(u_{w_i} v_{w_j}\right)}$.

Recent studies \cite{Meng2019SphericalTE,Meng2020DiscriminativeTM} also observe the importance of modeling the documents where a word appears in, since words in similar documents share topical coherence. We denote document embedding as $d$ and maximize the log probability of predicting the correct document that a word belongs to.
\begin{equation}
    \label{eq:local}
    \begin{aligned}
        L_d &= -\sum_{d\in D} \sum_{1\leq i\leq |d|} \log P(d\mid w_i)
    \end{aligned}
\end{equation}
where $P(d\mid w_i) \propto \exp{\left(u_{w_i} u_d\right)}$.


Since we want to regularize the embedding space to be discriminative among the concepts in the expanded taxonomy, we wish to form topical clusters in the embedding space where each concept embedding is surrounded by its representative terms.
We use an iterative approach to gradually grab distinctive words at each epoch. Specifically, we add one distinctive word to each concept cluster $C_e$ at each epoch to avoid semantic drift.
We then enforce the proximity between concept embedding and their clusters by 
\begin{align}
\label{eq:reg}
L_{prox} &= \sum_{e \in \mathcal{T}} \sum_{w \in \mathcal{C}_e} \log P(e \mid w)
\end{align}
where $P(e \mid w) \propto \exp{\left(u_{w_i} u_e\right)}$

The overall training objective is a weighted sum of the above terms.
\begin{align}
\label{eq:obj}
 L=L_l + \lambda_d L_d + \lambda_{p} L_{prox} 
\end{align}




\subsubsection{Topic and Relation aware Subtopic Finding}\label{sec:depth}
To find subtopics for existing concepts in the seed taxonomy, we apply two constraints for generating potential candidates for subtopics: (1) Topical constraints: candidates should belong to the topic cluster $C_e$ of that concept $e$; and (2) relational constraints: candidates should bear user-interested relation with the concept. The two constraints can be applied by using the learned relation classifier and concept embedding.
However, directly using the few-shot relation classifier can still include noisy and non-consistent terms, thus we carry out a co-clustering method to further filter out those noisy terms. 

An example is shown in Table \ref{tab:nocon}, where we use a Topic-Type table to organize the valid terms from the relation classifier. Valid terms are divided in columns by semantic meaning and in rows by semantic type (e.g., \textit{food}, \textit{cooking style} or \textit{sauces}). It is easily observable that the fourth subtopic of ``pieces, slices'' is an outlier sharing little type similarity with other subtopics. Thus we apply co-clustering method to retain those subtopics sharing similar semantic type distribution. 

{
	\setlength{\tabcolsep}{3pt}
	\begin{table}[htb]
		\centering
		\scalebox{0.85}{
			\begin{tabular}{|c|c|c|c|c|}
				\hline
				\multicolumn{2}{|c|} {\textbf{Beef}} & \textbf{Pork} & \multicolumn{2}{c|} {\textbf{Bread}} \\
				\hline
				sliced beef & \makecell{sirloin, rare beef} & roasted pork &  & \makecell{flat bread, wheat}\\ 
				\hline
				\makecell{stewed} & & & & toasted \\
				\hline
				& & & \makecell{pieces,  slices} & \\
				\hline
				& black pepper & spicy sauce & & buttery\\
				\hline
		\end{tabular}}
		\caption{An example of a Topic-Type table.}
		\label{tab:nocon}
	\end{table}
}


\smallskip
\noindent \textbf{Topic-Type Matrix Creation.}~
We construct an indicative (0/1) Topic-Type matrix to represent the joint distribution of subtopics and types of candidates from the Topic-Type table as shown in Table \ref{tab:nocon}, and the table is created by the following process: 
The topic-wise clustering is done by affinity propagation (AP) clustering \cite{Frey2007ClusteringBP} in the discriminative embedding space trained by concept learning, where the concepts are separated away from each other to avoid overlapping. 
The type-wise clustering is conducted by AP on the average BERT embedding space: We first retrieve the contextualized embedding of each candidate mention using the last layer output of BERT, and then average over the mentions to get the embedding for each candidate. 

\smallskip
\noindent \textbf{Co-clustering of the Matrix.}~
Finally, to extract high quality sub\-topics, we apply co-clustering \cite{Kluger2003SpectralBO} on the indicative Topic-Type matrix $M$ and define a consistency score for each cluster.
\begin{align}
\text{Consistency}(\text{Cluster}_k) = \frac{\sum_{ \text{row label[i]==k}}\sum_{\text{col. label[j]==k}} M_{i,j} }{ \sum_{ \text{row label[i]==k}}\sum_{\text{col. label[j]==k}} 1 } 
\end{align}
If the consistency score of a cluster is high, then the cluster consists of multiple subtopics that share similar semantic types. We retain high quality subtopics by setting a threshold for the consistency score.

\subsection{Overall Algorithm}
We summarize the overall algorithm of seed-guided topical taxonomy construction in Algorithm \ref{alg:overall}.

\begin{algorithm}[ht]
\caption{Seed-guided Topical Taxonomy Construction.}
\label{alg:overall}
\KwIn{
A text corpus $\mathcal{D}$; a given taxonomy $\mathcal{T}^0$ consisting of nodes $\{e_{i}\}|_{i=1}^{n}$ and edges ${\la p_i,c_i\ra}|_{i=1}^{m}$.
}
\KwOut{A more complete taxonomy $\mathcal{T}$ with each node \textbf{$e$} represented by a cluster of terms.}
Initialize relation training sample list $\mathcal{S}$\;
\For{$i \gets 1$ to $m$} {
Extract sentences $S_{p_i,c_i}$ where $ p_i$ and $c_i$ co-occur\;
$\mathcal{S} \gets S_{p_i,c_i}$\;
}
Train the relation classifier $F$ according to Section \ref{sec:rel}\;
$R \gets$ root nodes discovered by Section \ref{sec:width}\;
Initialize new first-layer topic candidates $e_{\text{new}}$ by co-occurred terms of $r \in R$\;
$\text{Score}(R\rightarrow e_{\text{new}}) \gets $ Equation \ref{eq:extraScore} \;
$\mathcal{B}_{R} \gets \mathcal{B}_{R} \cup \{e_{\text{new}} \text{ if } \text{Score}(R\rightarrow e_{\text{new}}) > \gamma \} $ \; \Comment{New first-layer topics found are attached to root node.}\;
\For{internal nodes $e$ in $\mathcal{T}$} {
$\textbf{B} \gets$ Candidate terms extraction by Section \ref{sec:cand} \;
$\mathcal{B}_e \gets \mathcal{B}_e \cup \textbf{B}$\; 
}
Train a joint embedding space of words and concepts\;
Extract topical words $C_e$ for internal nodes $e \in \mathcal{T}$ \;
Subtopic Finding by Section \ref{sec:depth}\;
Return topical taxonomy $\mathcal{T}$\;
\end{algorithm}

\section{Experiments and Results}
\subsection{Experiment Setup}

\smallskip
\noindent \textbf{Datasets.}~
Our experiments are conducted on two large real-world datasets: (1) \textbf{DBLP} contains around 157 thousand abstracts from publications in the field of computer science. For preprocessing, we use AutoPhrase \cite{Shang2017AutomatedPM} to extract meaningful phrases to serve as our vocabulary, we further discard infrequent terms occuring less than 50 times, resulting in 16650 terms. (2) \textbf{Yelp} is collected from the recent released \textit{Yelp Dataset Challenge}\footnote{https://www.yelp.com/dataset/challenge}, containing around 1.08 million restaurant reviews. Similarly, we extract meaningful phrases and remove infrequent terms, resulting in 14619 terms\footnote{Our code and data are available at \url{https://github.com/teapot123/CoRel}}. 

\smallskip
\noindent \textbf{Hyperparameter setting.}~
For our relation classifier, the hyperparameter is set to be: batch size = 16, training epochs = 5, model: Bert-Base (12 layers, 768 hidden size, 12 heads). When training the relation classifier, we make a 90/10 training/validation split on training samples. In our relation transferring process, we set the threshold for relation score in Eqs.~\eqref{eq:score} and \eqref{eq:extraScore} to 0.7, and the threshold for KL divergence $\delta$ to 0.5. 
For our concept learning module, we set the following hyperparameters for embedding training process: embedding dimension = 100, local context window size = 5, $\lambda_d$ = 1.5, and $\lambda_{p}$ = 1.0. The threshold for Cluster Consistency is 0.5. We use the same hyperparameters for both datasets.



\smallskip
\noindent \textbf{Compared Methods.}~
We compare \corel to several previous corpus-based taxonomy construction algorithms. To the best of our knowledge, there is no seed-guided topical taxonomy construction methods where each node is represented by a cluster of words, so we also compare \corel with some unsupervised methods. 

\begin{itemize}[leftmargin=*]
	\item Hi-Expan \cite{Shen2018HiExpanTT} + Concept Learning: Hi-Expan is a seed-guided instance-based taxonomy construction algorithm, which has the same input as our setting and constructs the taxonomy structure by set expansion and word analogy. Since its output node is represented by single word, we apply our concept learning module to enrich each node with a cluster of words.
	\item TaxoGen \cite{Zhang2018TaxoGenUT}: An unsupervised topical taxonomy construction method. It uses spherical clustering and local-corpus embedding to discover fine-grained topics represented by clusters of words.
	\item HLDA \cite{Blei2003HierarchicalTM}: A non-parametric hierarchical topic model. It models the generation of documents in a corpus as sampling words from the paths when moving from the root node to a leaf node. Thus we can take each node as a topic.
	\item HPAM \cite{Mimno2007MixturesOH}: A state-of-the-art hierarchical topic model which requires a pre-defined number of topics and outputs topics at different levels.
\end{itemize}

\subsection{Qualitative Results}
In this section we show the topical taxonomy generated by \corel on both datasets. We further exhibit the effectiveness of our concept and relation learning modules by comparing with baseline methods.

\begin{figure*}[htb]
    \centering 
\begin{subfigure}{0.25\textwidth}
  \includegraphics[width=\linewidth]{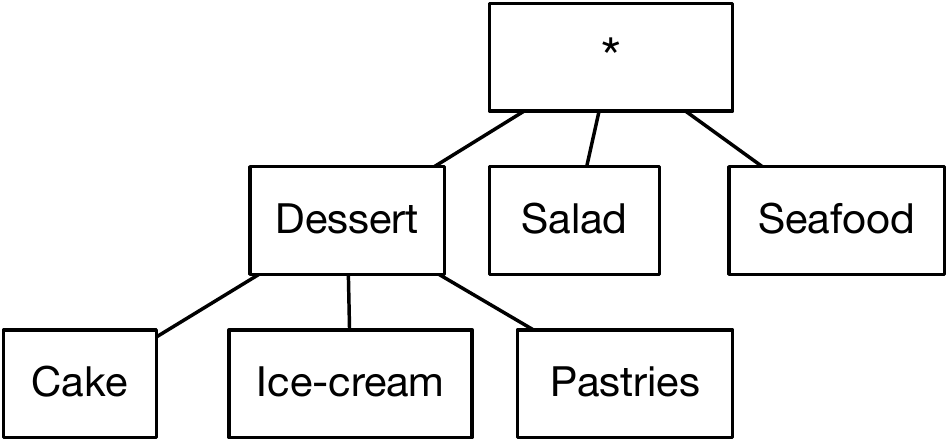}
  \caption{Seed of Yelp}
  \label{fig:yelp_in}
\end{subfigure}\hfil 
\begin{subfigure}{0.7\textwidth}
  \includegraphics[width=\linewidth]{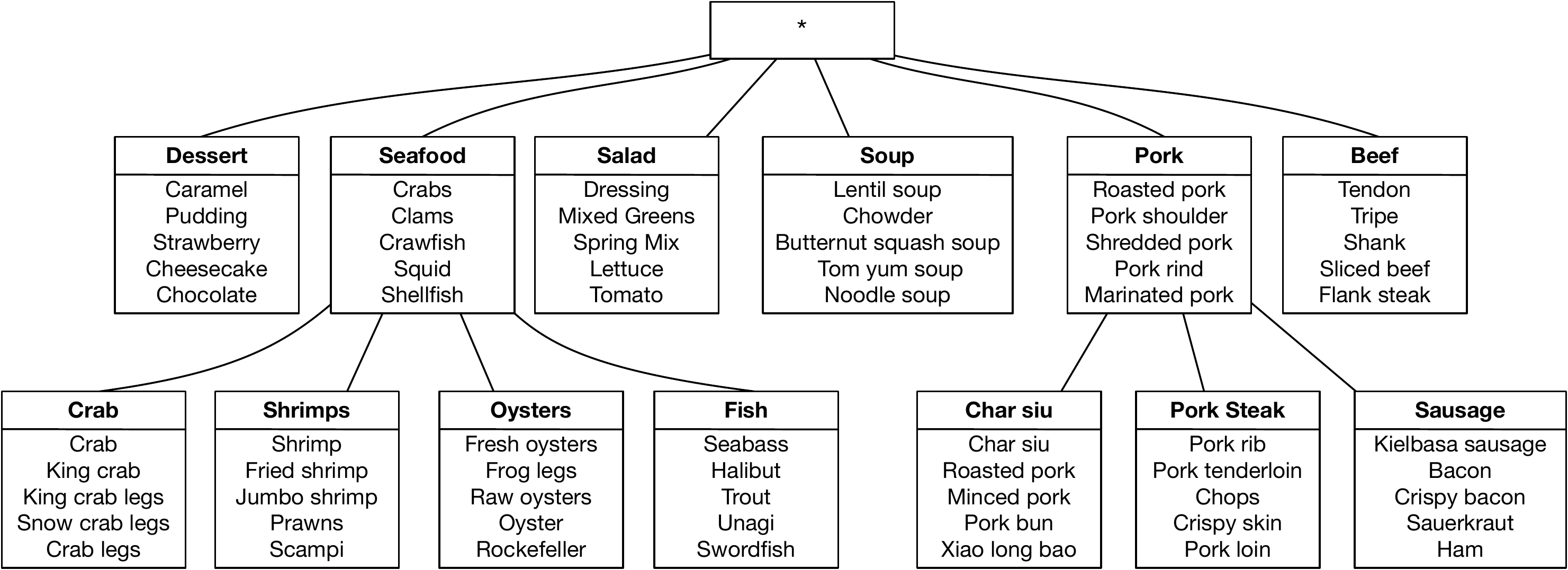}
  \caption{Parts of the taxonomy generated by \corel on the Yelp dataset}
  \label{fig:yelp_out}
\end{subfigure}
\medskip
\begin{subfigure}{0.8\textwidth}
  \includegraphics[width=\linewidth]{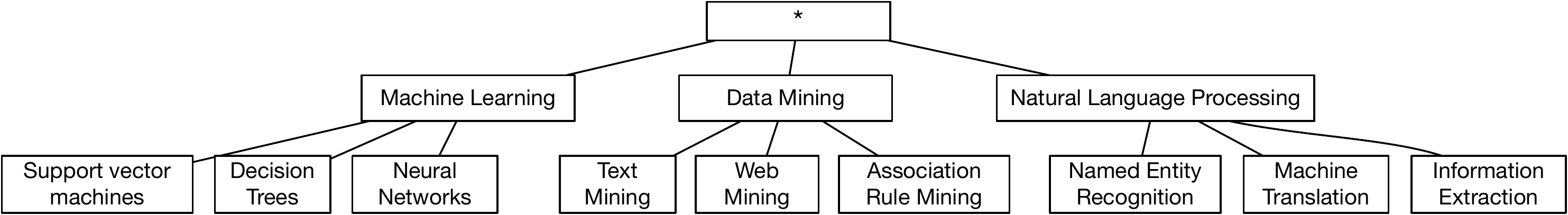}
  \caption{Seed of DBLP}
  \label{fig:dblp_in}
\end{subfigure}
\medskip
\begin{subfigure}{\textwidth}
  \includegraphics[width=\linewidth]{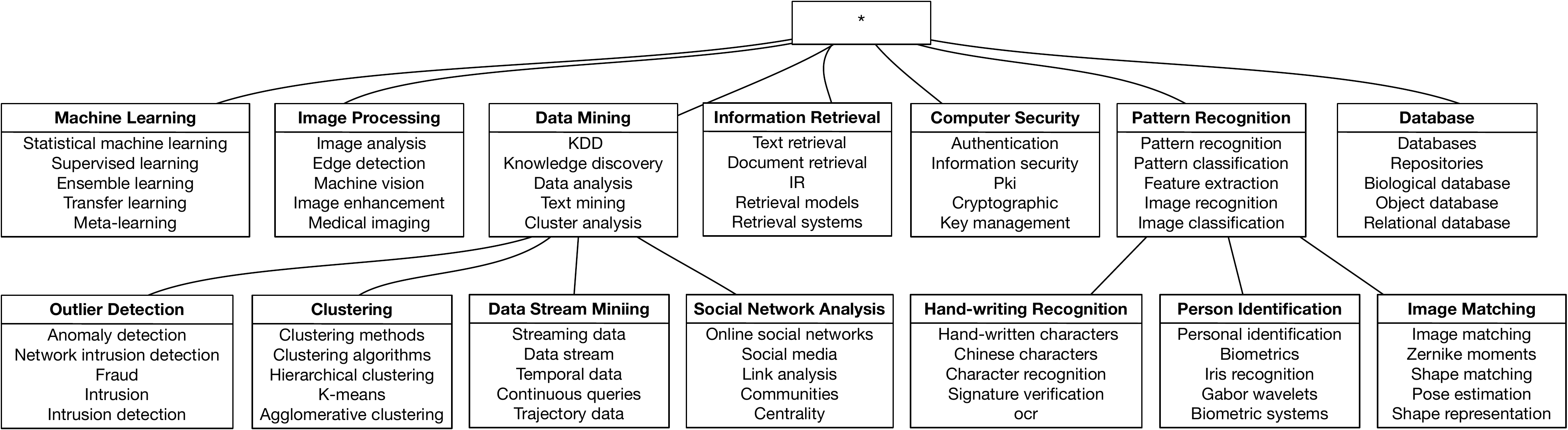}
  \caption{Parts of the taxonomy generated by \corel on the DBLP dataset}
  \label{fig:dblp_out}
\end{subfigure}
\caption{Input and part of output taxonomy generated by \corel on \textbf{DBLP} and \textbf{Yelp}.}
\vspace{-0.3cm}
\label{fig:inout}
\end{figure*}

\smallskip
\noindent \textbf{Our Topical Taxonomy.}~
Figure \ref{fig:inout} shows the input seed taxonomy and parts of the topical taxonomy generated by \corel on both datasets. 
For the \textbf{Yelp} dataset, we use minimal user input by only giving child nodes for one input topic to test the robustness of our method. The output in Figure \ref{fig:yelp_out} shows that we can find new food types such as ``soup'', ``pork'' and ``beef''. For subtopic finding, we can also distinguish between various western and eastern cooking style of pork. The word clusters for each topic/subtopic are obtained by the concept learning module trained on the corpus and the whole taxonomy. For \textbf{DBLP}, we use the same input seed as Hi-Expan \cite{Shen2018HiExpanTT}. We show that \corel successfully finds various computer science fields in Figure \ref{fig:dblp_out} other than user's input, such as ``information retrieval'' and ``pattern recognition''. \corel is also capable of finding separate fields for seed and new research areas found.

\smallskip
\noindent \textbf{Subtopic Quality.}~
We exhibit the effectiveness of our relation learning module by comparing the subtopics we found with those of HiExpan (seed-guided tree expansion baseline). We randomly choose the common topics shared by both methods, and show all subtopics found by each of them in Table \ref{tab:subtopic}. The bold ones are seeds from the input taxonomy, while the wrong subtopics are marked by ($\times$). Since generating synonyms or included concepts at the same level harms the overall quality of a taxonomy, we mark these terms as redundant (\redund). 
For example, under the topic of ``\textbf{DBLP}-Machine Learning'', HiExpan generates lots of synonyms and subconcepts for ``neural networks'' at the same level, such as ``artificial neural networks'' and ``multilayer perceptron''. These terms should be formed in the topic of ``neural networks'' instead of its sibling nodes. Our design of concept learning puts these terms into the distinctive term clusters of corresponding topics, thus avoiding such pitfalls. Under the topic of ``\textbf{Yelp}-Seafood'', we show that simply using word analogy is not robust in capturing parallel relations in the global embedding space, and it is essential to utilize more advance text representations and operations upon them as in our relation learning module.

\begin{table*}
\centering
\caption{Comparison of subtopics found under the same topics.}
\label{tab:subtopic}
\scalebox{0.88}{
\begin{tabular}{|c|c|c|}
\hline
Topics & Method & All Subtopics found\\
\hline
\multirow{2}{*}{\makecell{ \textbf{DBLP}-Machine \\ Learning}} & Hi-Expan
& \makecell{ 
\textbf{Support vector machine}, \textbf{Decision trees}, \textbf{Neural networks}, Regression models, Genetic algorithm, \\
Naive Bayes, Classification, Random forest, Markov random field,  Nearest Neighbor, Unsupervised learning,\\ Artificial neural networks (\redund), Multilayer perceptron (\redund), Support vector regression (\redund), \\
Conditional random fields (\redund), hidden markov models (\redund), Radial basis function(\redund), Self-organizing map (\redund), \\
Recurrent neural networks (\redund), Extreme learning machine(\redund), \wrong{Particle swarm optimization},}\\
\cline{2-3}
& \corel 
& \makecell{ 
\textbf{Support vector machine}, \textbf{Decision trees}, \textbf{Neural networks}, Regression, Genetic algorithm, \\ 
Bayesian networks, Classification, Random forest, Inductive logic programming, Reinforcement learning,\\
Active learning, Boosting algorithms, Transfer learning, \wrong{object recognition},  \wrong{Text classification}}\\
\hline
\multirow{2}{*}{\makecell{ \textbf{DBLP}-Data \\ Mining}} & Hi-Expan
& \makecell{ \textbf{association rule mining}, \textbf{text mining}, \textbf{web mining}, outlier detection, 
anomaly detection, \\
spectral clustering, social network analysis, density estimation,(\redund) association rules (\redund)}\\
\cline{2-3}
& \corel 
& \makecell{ \textbf{association rule mining}, \textbf{text mining}, \textbf{web mining}, outlier detection, 
anomaly detection,\\
clustering algorithms, social network analysis, data stream mining, data visualization,\\
online analytical processing, rule discovery, predictive modeling, sequence analysis (\redund)}\\
\hline
\multirow{2}{*}{\makecell{ \textbf{Yelp}-Dessert}} & Hi-Expan
& \makecell{\textbf{cake}, \textbf{ice cream}, \textbf{pastries}, latte, cream, gelato, \wrong{sauce},   cheesecake (\redund), \wrong{taste}, pancake (\redund)}\\
\cline{2-3}
& \corel 
& \makecell{\textbf{cake}, \textbf{ice cream}, \textbf{pastries}, milk tea, cream, fruit, juice, \wrong{cooked}, \wrong{sugar}}\\
\hline
\multirow{2}{*}{\makecell{ \textbf{Yelp}-Seafood}} & Hi-Expan
& \makecell{fish, shrimps, salmon (\redund), \wrong{meat}, \wrong{chicken}, \wrong{beef}, \wrong{steak},  \wrong{pork}, \wrong{rice}}\\
\cline{2-3}
& \corel 
& \makecell{fish, shrimps, crab, scallop, oysters, mussel, camarones (\redund), \wrong{soy sauce}, \wrong{pho}, \wrong{low mein} }\\
	    
\hline

\end{tabular}
}
\end{table*}

\smallskip
\noindent \textbf{Concept clusters.}~
We wish to evaluate how well our concept learning module forms meaningful clusters. We compare with TaxoGen, HLDA and HPAM that output hierarchical clusters of terms. Since they do not need user-given seeds as supervision, and generate many irrelevant topics, we set the number of topics for TaxoGen and HPAM to 10 at each layer, and use the default setting for HLDA. Then we manually pick out common topics/subtopics found by different algorithms. We list the centermost 5 terms for \corel topics/subtopics, and the 5 terms with top probability of TaxoGen, HLDA and HPAM in Table \ref{tab:quality}. We observe that without user-given seeds, topics from TaxoGen, HLDA and HPAM include mixtures of cross-concept terms or irrelevant terms, while our concept learning module is able to find coherent and distinctive terms for each concept.

\setlength{\tabcolsep}{1pt}
\begin{table}[ht]
	\small
	\centering
	\caption{Topic clusters generated by different methods.}
	\label{tab:quality}
	\scalebox{0.95}{
		\begin{tabular}{c|c|c|c}
			\toprule
            Met.
            & \makecell{DBLP \\ Recommender System}
            & \makecell{DBLP \\ Image Matching}
            & \makecell{Yelp \\ Beef} \\
			\midrule
			
			\multirow{5}{*}{\makecell{HLDA}} 
			& recommended items & illumination variation &  reuben    \\
			& recommendation framework & \wrong{relevant document} & \wrong{Don Juan}    \\
			& \wrong{sentimental analysis} & affine distortion  & \wrong{cutter}    \\
			& \wrong{source entropy} & BNMTF  & corned beef    \\
			& \wrong{customer} & \wrong{phase diagram} &  turnover beef    \\
			\midrule
			
			\multirow{5}{*}{\makecell{HPAM}} 
			& ranking & face recognition  & BBQ    \\
			& \wrong{web page} & \wrong{image} & brisket    \\
			& \wrong{propose} & \wrong{video} &  ribs    \\
			& \wrong{different} & \wrong{detection} & \wrong{meat}    \\
			& recommendation & \wrong{segmentation} & good    \\
			\midrule
			
			\multirow{5}{*}{\makecell{TG}}
			& \wrong{linked data} & fisher criterion & wellington  \\
			& \wrong{social network analysis} & face verification &  wagyu beef \\
			& recommendation systems & \wrong{decision fusion} & dry aged  \\
			& user interests & \wrong{classifier design}  & \wrong{walleye}  \\
			& user feedback & \wrong{discriminative power}  & \wrong{red meat}  \\
			\midrule
			
			
			\multirow{5}{*}{\makecell{\corel}}
			& recommender systems & image matching & tendon  \\
			& collaborative filtering & zernike moments & tripe  \\
			& recommendation & shape matching & beef ball \\
			& user preferences & pose estimation  & flank  \\
			& user rating & feature point & rare beef  \\

			\bottomrule
		\end{tabular}
	}
	\vspace*{-1em}
\end{table}

\subsection{Quantitative Results}
In this section, we quantitatively evaluate the quality of the taxonomies constructed by different methods.

\smallskip
\noindent \textbf{Evaluation Metrics.}~
The evaluation of the quality of a topical taxonomy is a challenging task since there are different aspects to be considered, and it is hard to construct a gold standard taxonomy that contains all the correct child nodes under each parent node. Following \cite{Shen2018HiExpanTT,Zhang2018TaxoGenUT}, we propose three evaluation metrics: \textit{Term Coherency}, \textit{Relation F1} and \textit{Sibling Distinctiveness} in this study.

\begin{itemize}[leftmargin=*]
\item \textbf{Term Coherency (TC)} measures the semantic coherence of words in a topic.
\item \textbf{Relation F1} measures the portions of correct parent-children pairs in a taxonomy that preserve user-interested relation.
\item \textbf{Sibling Distinctiveness (SD)} measures how well the topics are distinctive from their siblings.
\end{itemize}

We calculate the three metrics as follows. For \textit{TC}, we recruited 5 Computer Science students to judge the results. Specifically, we extract the top 10 representative words under each topic, ask the evaluators to divide these words into different clusters by concepts and compute the size of the cluster with the most words, thus a mixture of terms from different concepts scores lower. Then we take the mean of the results given by all the evaluators as the \textit{TC} of this topic, and average the \textit{TC} of all the topics in a taxonomy.

For \textit{Relation F1}, we show the evaluators all the parent-children pairs of topics in a taxonomy and ask them to judge independently whether each pair truly holds user-interested relation. Then we use majority votes to label the pairs and use all the true parent-children pairs from different methods to construct a gold standard taxonomy. Since each topic is represented by a cluster of words, for simplicity, we consider two clusters as the same if they share the same concept.
The \textit{Relation F1} is computed as follow:
{\small
\begin{equation}
\begin{aligned}
    P_{r} &= \frac{|is\_ancestor_{pred}|\bigcap|is\_ancestor_{gold}|}{|is\_ancestor_{pred}|},\\
    R_{r} &= \frac{|is\_ancestor_{pred}|\bigcap|is\_ancestor_{gold}|}{|is\_ancestor_{gold}|},\\
    F1_{r} &= \frac{2P_r*R_r}{P_r + R_r}
\end{aligned}
\end{equation}
}
where $P_r$, $R_r$ and $F1_r$ denote the \textit{Relation Precision}, \textit{Relation Recall} and \textit{Relation F1}, respectively.

Finally, we calculate \textit{Sibling Distinctiveness (SD)} as follows: we compute the similarity between a topic cluster $C_i$ and each of its sibling topics $C_j$ by Jaccard index \cite{Ville2001IntroductionTD}. Then we calculate \textit{SD} of $C_i$ as $1$ minus the largest similarity score among all $C_j$. A larger \textit{SD} means the sibling topics sharing a common parent are truly separate from each other.

\smallskip
\noindent \textbf{Evaluation Results.}~
Table \ref{tab:quantity} shows the \textit{Term Coherency (TC)}, \textit{Relation F1}, and \textit{Sibling Distinctiveness (SD)} of different methods. For unsupervised baselines, we only take relevant topics (topics with more than half terms belonging to \textit{food} or \textit{research areas}) into account. 
Overall, weakly-supervised methods (Hi-Expan and \corel) outperform unsupervised methods by a large margin, which shows the constructed taxonomies are well guided by the user given seeds.
We can see that \corel achieves the best performance under all evaluation metrics, especially in terms of the \textit{Relation F1}, showing that \corel is able to find related terms for each concept and retain ones holding certain relations with current topics in relation transferring module. For \textit{TC}, \corel also significantly outperforms HLDA, HPAM and TaxoGen, which model the intrinsic distribution of terms in documents or corpus, and might generate topics as mixtures of terms relevant but not distinctive of user-interested topics. They also have inferior performance in \textit{SD} since they do not enforce distinctiveness when forming topics. 
On the other hand, though HiExpan $+$ Concept Learning only achieves slightly worse or equal results on \textit{TC} and \textit{SD} compared with ours, HiExpan itself only outputs an instance taxonomy and cannot generate topics for each concept node. We enhance it by our own concept learning module to extract distinctive terms for each node. This further demonstrates the effectiveness of both of our modules.

\begin{table*}[ht]
	\centering
	\small
	\caption{Quantitative evaluation on topical taxonomies.}
	\label{tab:quantity}
	\scalebox{1.0}{
	\setlength{\tabcolsep}{8pt}
		\begin{tabular}{c|ccccc|ccccc}
			\toprule
			\multirow{2}{*}{Methods} &
			\multicolumn{5}{c|}{\textbf{DBLP}} & \multicolumn{5}{c}{\textbf{Yelp}} \\
			& \ \ TC\ \  & SD & $\text{Precision}_r$ & $\text{Recall}_r$ & $\text{F1-score}_r$
			& TC & SD & $\text{Precision}_r$ & $\text{Recall}_r$ & $\text{F1-score}_r$ \\
			\midrule
			HLDA 
			& 0.582 & 0.981 & 0.188 & 0.577 & 0.283 
			& 0.517 & 0.991 & 0.135 & 0.387 & 0.200\\
			HPAM 
			& 0.557 & 0.905 & 0.362 & 0.538 & 0.433
			& 0.687 & 0.898 & 0.173 & 0.615 & 0.271\\
			TaxoGen 
			& 0.720 & 0.979 & 0.450 & 0.429 & 0.439  
			& 0.563 & 0.965 & 0.267 & 0.381 & 0.314\\
			Hi-Expan $+$ CoL.
			& 0.819 & 0.996 & 0.676 & 0.532 & 0.595 
			& 0.815 & \textbf{1.000} & 0.429 & 0.677 & 0.525 \\
			\corel 
			& \textbf{0.855} & \textbf{1.000} & \textbf{0.730} & \textbf{0.607} & \textbf{0.663} 
			& \textbf{0.825} & \textbf{1.000} & \textbf{0.564} & \textbf{0.710} & \textbf{0.629} \\
			\bottomrule
		\end{tabular}
	}
\end{table*}

\section{Conclusions and Future Work}
In this paper we explore the problem of seed-guided topical taxonomy construction. Our proposed framework \corel completes the taxonomy structure by a relation transferring module and enriches the semantics of concept nodes by a concept learning module. The relation transferring module learns the user-interested relation preserved in seed parent-child pairs, then transfers it along multiple paths to expand the taxonomy in width and depth. The concept learning module finds discriminative topical clusters for each concept in the process of jointly embedding concepts and words. Extensive experiments show that both modules work effectively in generating a high-quality topical taxonomy based on user-given seeds.

For future work, it is interesting to study how we can generate multi-faceted taxonomy automatically, so that each concept node is described by terms from different aspects (e.g., \textit{ingredients} and \textit{cooking style} for foods). Though these terms can be captured by our concept learning module, how to recognize them and organize them into meaningful clusters remains challenging and worth exploring.

\begin{acks}
Research was sponsored in part by US DARPA KAIROS Program No. FA8750-19-2-1004 and SocialSim Program No.  W911NF-17-C-0099, National Science Foundation IIS 16-18481, IIS 17-04532, and IIS 17-41317, and DTRA HDTRA11810026. Any opinions, findings, and conclusions or recommendations expressed herein are those of the authors and should not be interpreted as necessarily representing the views, either expressed or implied, of DARPA or the U.S. Government. The U.S. Government is authorized to reproduce and distribute reprints for government purposes notwithstanding any copyright annotation hereon. 
We thank anonymous reviewers for valuable and insightful feedback.
\end{acks}

\bibliographystyle{ACM-Reference-Format}
\balance
\bibliography{acmart.bib}


%


\end{document}